\pdfoutput=1

\documentclass[11pt]{article}

\usepackage{acl}
\usepackage{hyperref}
\usepackage{times}
\usepackage{latexsym}
\usepackage{multirow}
\usepackage{booktabs}
\usepackage{microtype}
\usepackage{multirow}
\usepackage{latexsym}
\usepackage{booktabs}
\usepackage{courier}
\usepackage{amsmath}
\usepackage{booktabs} 
\usepackage{siunitx} 
\usepackage{subfig}
\usepackage{placeins}
\usepackage{tikz}
\usepackage{hyperref}

\usepackage[T1]{fontenc}

\usepackage[utf8]{inputenc}

\usepackage{microtype}

\usepackage{inconsolata}

\title{ISDS-NLP at SemEval-2024 Task 10: Transformer based \\ neural networks for emotion recognition in conversations}

\author{
Claudiu Creangă$^{2,3,}$, Liviu P. Dinu$^{1,3}$  \\ 
  $^1$ Faculty of Mathematics and Computer Science  \\
  $^2$ Interdisciplinary School of Doctoral Studies, 
  $^3$ HLT Research Center \\
  University of Bucharest, Romania\\
  \small
{\tt claudiu.creanga@s.unibuc.ro, ldinu@fmi.unibuc.ro}  \\
}
\vspace{2mm}

\begin{document}
\maketitle
\begin{abstract}

This paper outlines the approach of the ISDS-NLP team in the SemEval 2024 Task 10: Emotion Discovery and Reasoning its Flip in Conversation (EDiReF). For Subtask 1 we obtained a weighted F1 score of 0.43 and placed 12 in the leaderboard. We investigate two distinct approaches: Masked Language Modeling (MLM) and Causal Language Modeling (CLM). For MLM, we employ pre-trained BERT-like models in a multilingual setting, fine-tuning them with a classifier to predict emotions. Experiments with varying input lengths, classifier architectures, and fine-tuning strategies demonstrate the effectiveness of this approach. Additionally, we utilize Mistral 7B Instruct V0.2, a state-of-the-art model, applying zero-shot and few-shot prompting techniques. Our findings indicate that while Mistral shows promise, MLMs currently outperform them in sentence-level emotion classification.

\end{abstract}

\section{Introduction}

Task 10 from SemEval 2024 competition \cite{kumar2024semeval} addresses the complex challenge of identifying the emotions within dialogues (English and Hindi). This task comprises two primary objectives: firstly, assigning an emotion label to each utterance within a dialogue, and secondly, discerning the trigger utterance or utterances responsible for an emotion-flip within the dialogue \cite{KUMAR2022108112}. Emotions play a crucial role in human interaction and one can understand more from a text if one knows the underlying sentiment of the writer. In contexts where disagreements may arise, such as customer service platforms, virtual assistant chats or forums, identifying trigger utterances for emotion flips can help mediate conflicts and prevent escalation. A chatbot dealing with an angry customer would benefit from knowing how to speak in order to generate empathetic responses. If it knows that the chatbot's current sentence can trigger an emotion flip from neutral to anger, the chatbot should refine it, or if the emotion flip is from anger to joy, the chatbot should be more confident in such a response in the future. 

Both types of models we tried for Subtask 1 were based on transformers. The first one used BERT-like models and we achieved the best accuracy with them, while the second one is a state of the art causal model (Mistral, \cite{jiang2023mistral}) that was tested in zero-shot and few-shot settings with poorer results. 

Although in the first task our system worked well, placing 12th in the leaderboard, the other 2 tasks were much harder and we placed 14th on the second subtask. We believed that with a better strategy to prevent overfitting (like under or over-sampling), our system would have improved. Our code is open source and available to use on \href{https://github.com/ClaudiuCreanga/semeval-2024-task-10}{GitHub}.

\section{Background}

\begin{figure}[htbp]
    \centering
    \includegraphics[width=0.9\linewidth]{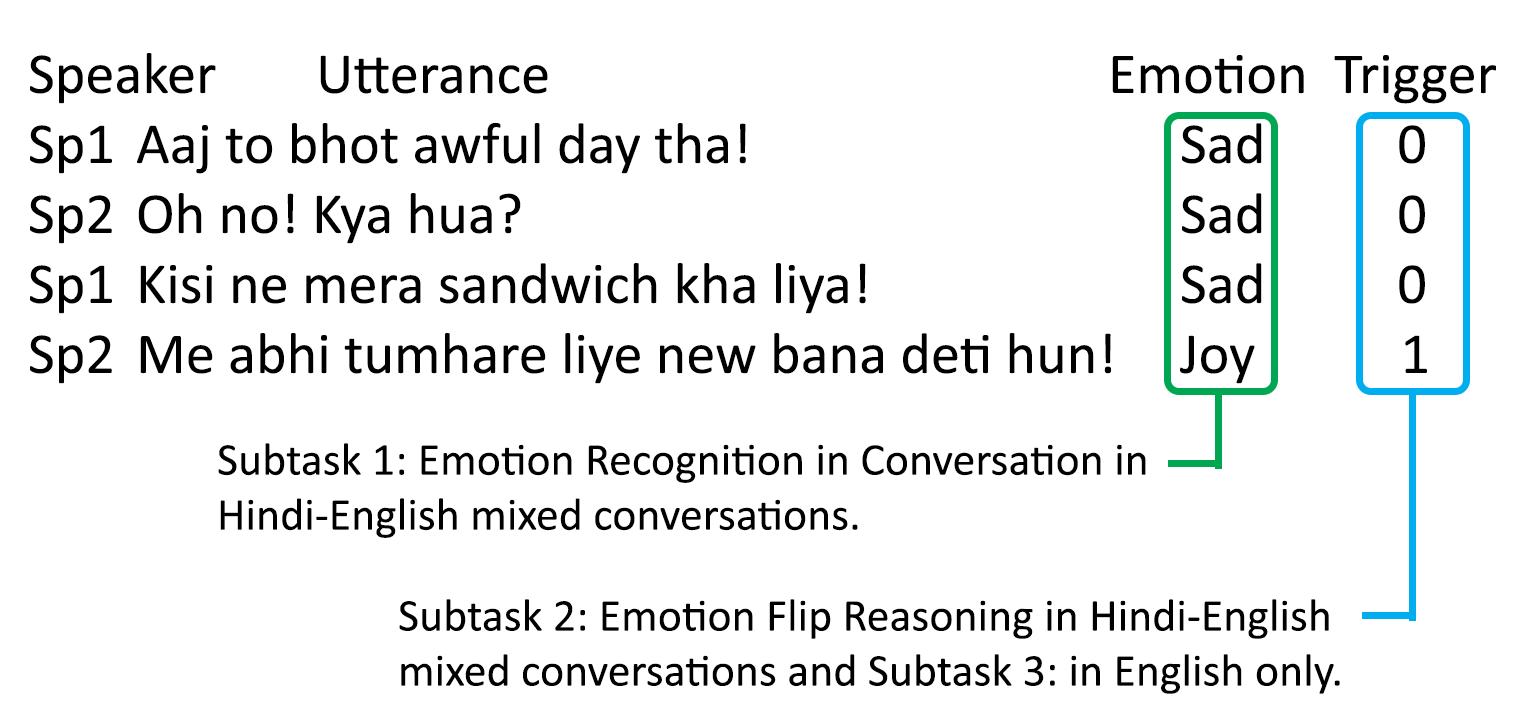}
    \caption{Three sub-tasks explained}
    \label{fig:three_subtasks}
\end{figure}

The competition had 3 subtasks explained in \autoref{fig:three_subtasks} and we participated in all of them with the best results on subtask 1 where we placed 12th with an F1 score of 0.43. 

\subsection{Dataset}

The dataset contains English and Hindi code-mixed conversations for Subtask 1 and 2 and English only conversations for Subtask 3 (\autoref{table:dataset_sizes}). The dataset is quite small, except for the training dataset for Subtask 2 and 3. If we were to combined them for Subtask 1, our F1 score would reach 0.97, but it wasn't allowed. This fact shows that with more data our model would do really well. The dataset is based on MELD, a known emotion recognition dataset, which was then augmented with triggers for the emotion-flip task. 

\begin{table}
  \caption{Datasets sizes used in this competition by tasks.}
  \label{table:dataset_sizes}
  \begin{tabular}{cccc}
    \toprule
      & Subtask 1 & Subtask 2 & Subtask 3 \\
    \midrule
    \texttt Train & 8506 & 98777 & 35000  \\
    \texttt Dev & 1354 & 7462 & 3522  \\
    \texttt Test & 1580 & 7690 & 8642  \\
    \bottomrule
  \end{tabular}
\end{table}

\begin{figure}[htbp]
    \centering
    \includegraphics[width=0.9\linewidth]{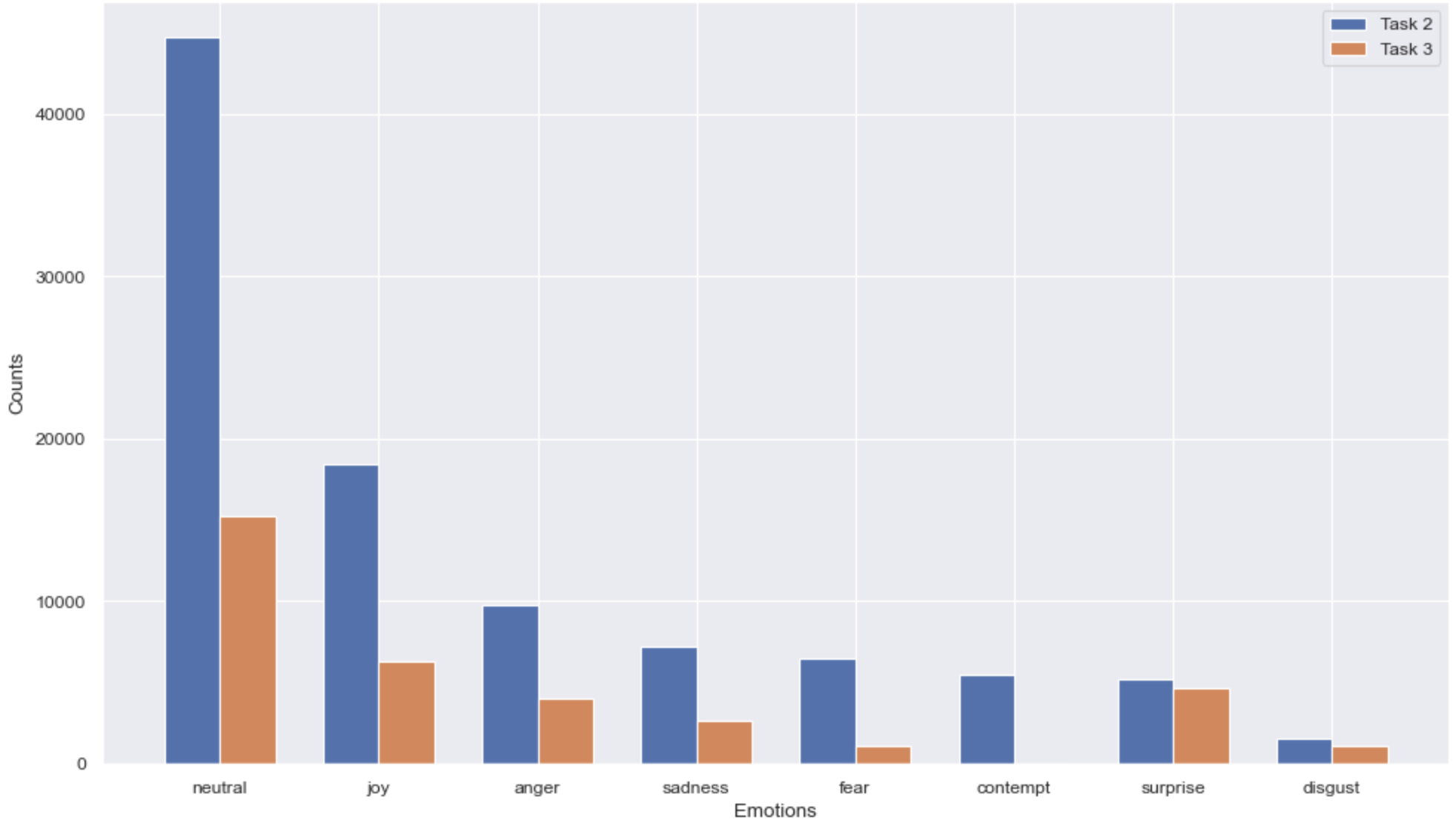}
    \caption{Emotion Distribution Comparison between Task 2 and Task 3}
    \label{fig:emotion_distribution}
\end{figure}

\begin{figure}[htbp]
    \centering
    \includegraphics[width=0.9\linewidth]{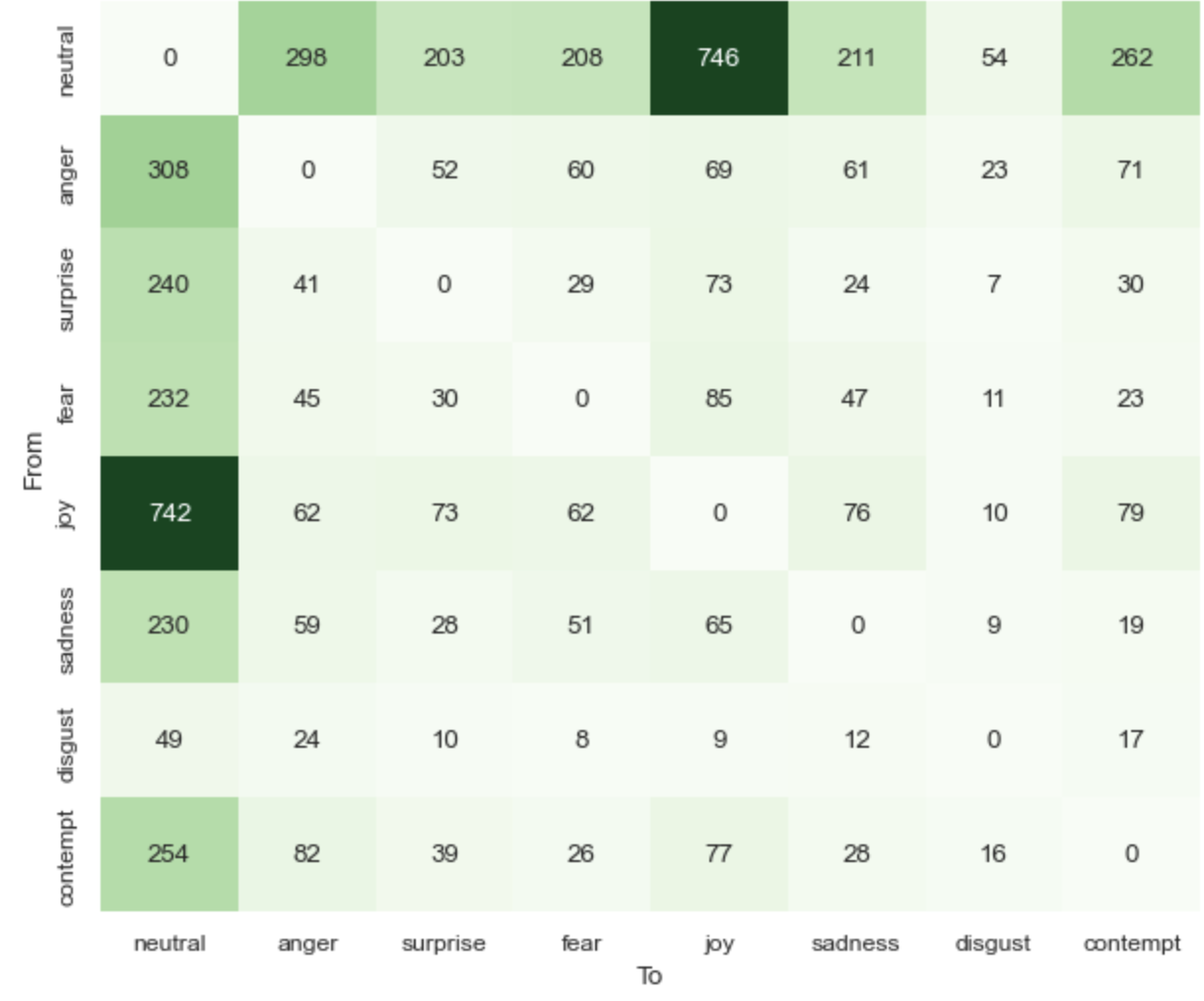}
    \caption{Task 2: Emotion-flip counts}
    \label{fig:flip_counts}
\end{figure}

There were 8 distinct emotions to predict: neutral, anger, surprise, fear, joy, sadness, disgust, and contempt. By far the most predominant emotion is neutral, followed by joy and anger (\autoref{fig:emotion_distribution}).  If we look at Subtask 2, most often the emotion flips are from neutral to joy or anger (\autoref{fig:flip_counts}).
 
\subsection{Previous Work}

Since the release of the first small datasets for emotion recognition in 1992 \cite{Ekman_1992}, the field has evolved substantially, marked by significant contributions from big companies in the form of extensive datasets \cite{demszky2020goemotions}. In the beginning, lexicon based methods were used in which there was a manually curated dictionary which associates words with specific emotions. The algorithm was simply picking the most expressed emotion according to the dictionary. This method had severe limitations because it was ignoring context, sentence structure and negations which can flip a sentiment. Today, state of the art models are based on transformer architecture and use either Masked Language Modelling (BERT based models \cite{Devlin}) or Causal Modelling (GPT \cite{Brown}) which can capture dependencies and nuances missed by word-level approaches.

While traditional emotion recognition tasks are well-established, research on emotion flip recognition is still in its early stages because it is a new task within the field of emotion analysis. Research \cite{Kumar} has found that a transformer based classifier with 6 encoder layer (EFR-TX) works well, obtaining an F1 score of 40 when trained on MELD-FR dataset and tested on IEMOCAP-FR dataset.

\section{System overview}

We tried two approaches, both of them based on transformer architecture: Masked Language Modelling and Casual Modelling. We chose these two architectures because of their recent successes in NLP.

\subsection{Masked Language Modelling}

\begin{figure}[htbp]
    \centering
    \includegraphics[width=1\linewidth]{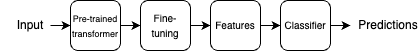}
    \caption{Model architecture}
    \label{fig:architecture}
\end{figure}

We used pre-trained BERT-like models in a multilingual setting so that it can tokenise Hindi sentences. These pre-trained models will give us the features  from sentences and then we pass them through a classifier which will do the prediction for each task \autoref{fig:architecture}. 

\subsubsection{Input}

\begin{figure}[htbp]
    \centering
    \includegraphics[width=1\linewidth]{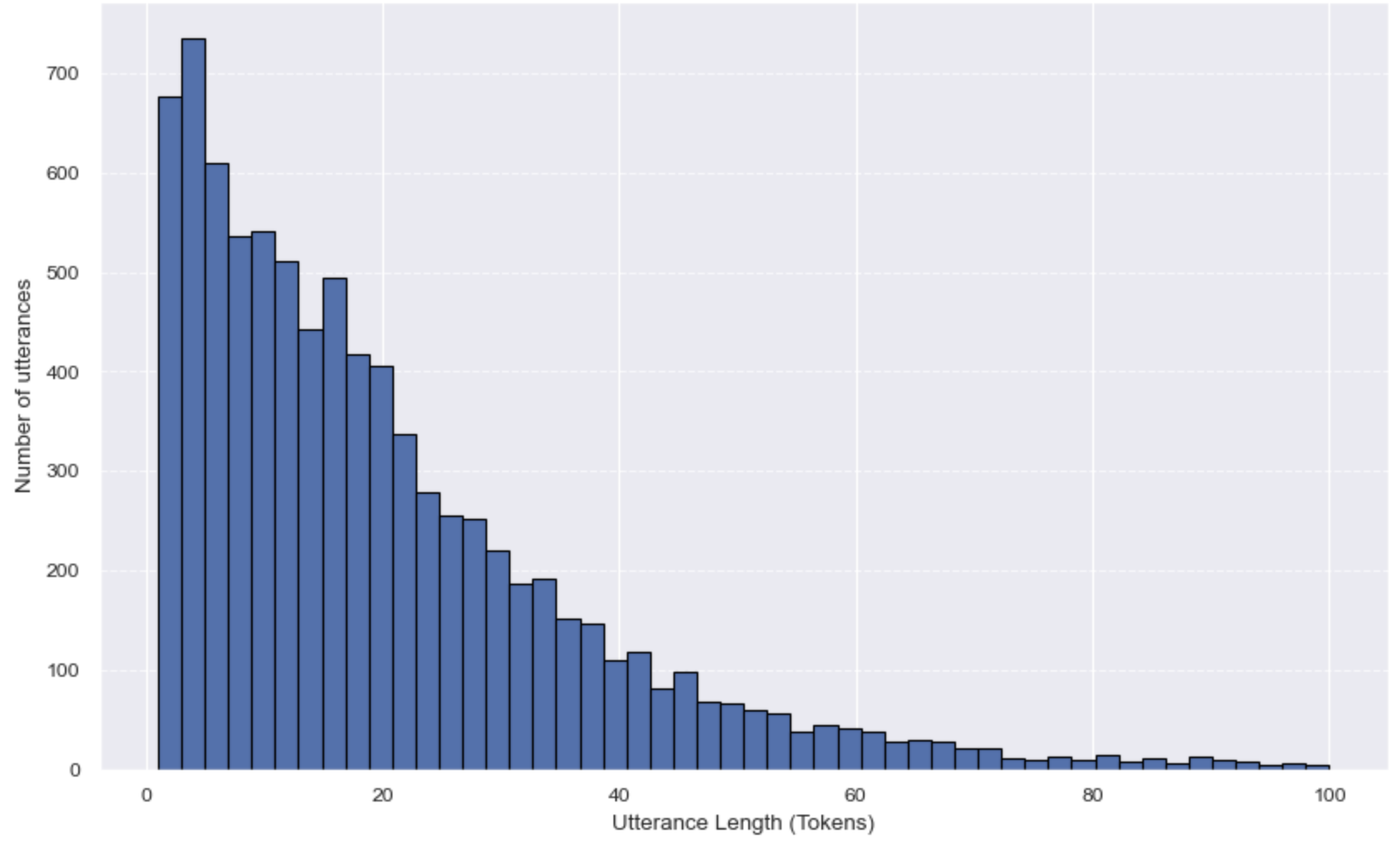}
    \caption{Distribution of utterances lengths.}
    \label{fig:tokens_lengths}
\end{figure}

\begin{figure}[htbp]
    \centering
    \includegraphics[width=1\linewidth]{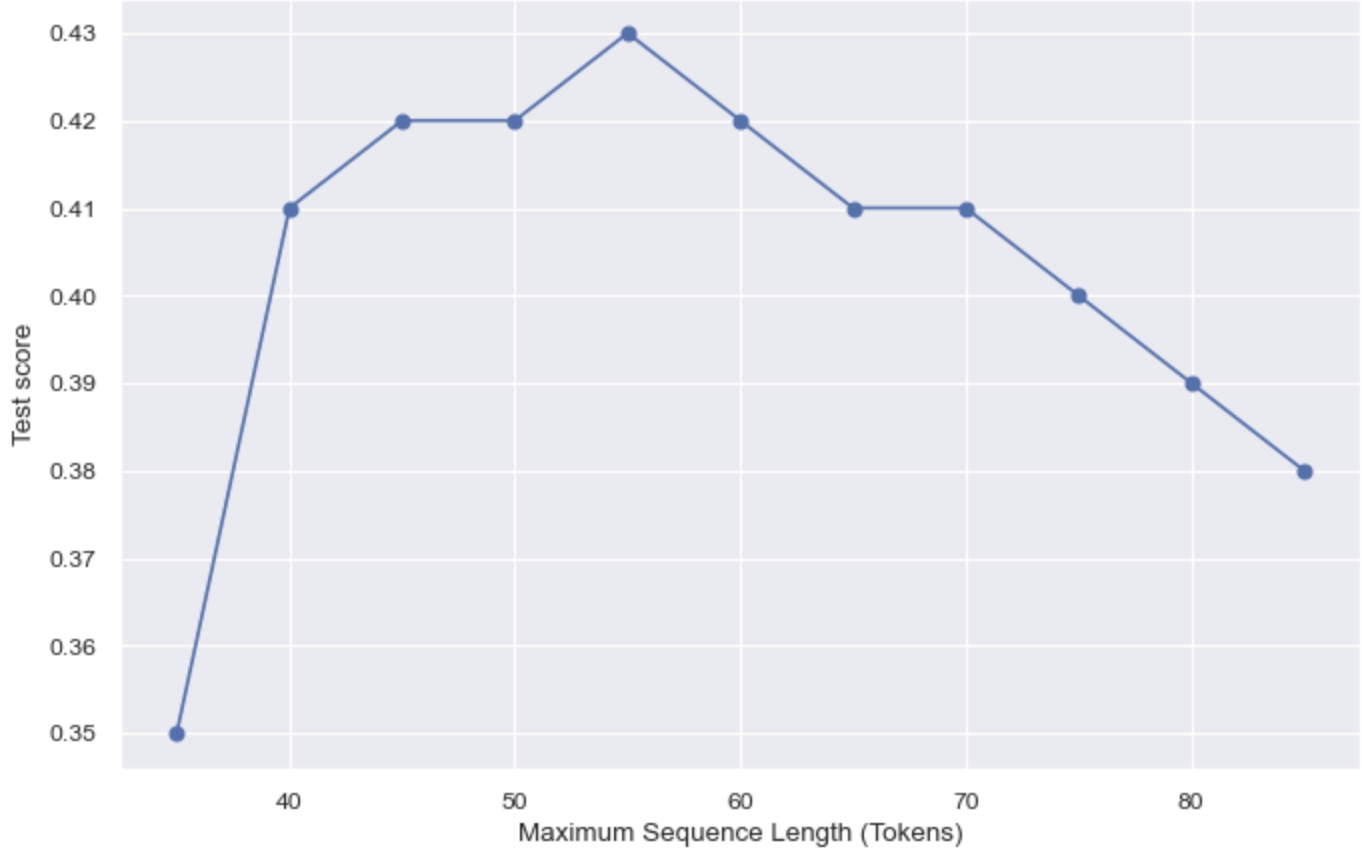}
    \caption{Best model score with different maximum utterances lengths.}
    \label{fig:score_tokens_lengths}
\end{figure}

Analysis of dialogue sentences reveals a predominantly short length, with a sharp decline in frequency after 30 tokens (see \autoref{fig:tokens_lengths}).  To optimize performance, various maximum sequence lengths were tested, with 55 tokens yielding the best results (\autoref{fig:score_tokens_lengths}). Data preprocessing, (such as lemmatization, removing punctuation or stopwords) didn't help the model learn better so we kept the input as is. Probably this is because punctuation and stopwords contain useful information that the models is able to learn.

\subsubsection{Output}

Selecting the optimal hidden state layer is crucial for leveraging the pre-trained model's results. Our experiments demonstrated that using the final layer's output yielded the strongest performance, with accuracy declining in earlier layers. For MLM-type models, the [CLS] token encodes the features, which is what we pass to our classification layer. 

Among various classifiers tested (\autoref{table:classifiers}), fully connected layers excelled, likely due to their ability to model complex, non-linear relationships. The top-performing model employed a fully connected layer with 0.5 dropout and a Softmax activation function.

\begin{table}
  \caption{Test scores of different classifiers.}
  \label{table:classifiers}
  \begin{tabular}{cccc}
    \toprule
     Classifier & Extra features & Score\\
    \midrule
    \texttt Fully Connected & Dropout(0.5) & 0.43  \\
    \texttt Fully Connected & Dropout(0.7) & 0.42  \\
    \texttt Fully Connected & Dropout(0.2) & 0.40  \\
    \texttt Fully Connected & - & 0.40  \\
    \texttt RandomForest & - & 0.23   \\
    \texttt LogisticRegression & - & 0.21  \\
    \texttt KNeighbors & - & 0.20   \\
    \bottomrule
  \end{tabular}
\end{table}

\subsubsection{Fine-tuning} \label{finetunning}

The large pre-trained language models we employed offer a robust foundation for understanding language in general. Through fine-tuning, we adapt them to the nuances of our emotion recognition task. Inspired by the strategy presented in  \cite{sun2020}, we initially train only the classifier with a larger learning rate (5e-5) and a warm-up period of 10,000 steps over 'k' epochs (we tried a range of 'k' from 1 to 10). Subsequently, we fine-tune both the classifier and the transformer's final layer using a smaller learning rate (2e-5). Our goal in freezing the transformer weights at first, and then training them with a reduced learning rate, is to minimize the risk of overfitting.

\subsection{Causal Modelling}

Given the success of generative models we also tried Mistral 7B Instruct V0.2  which is believed to be state of the art in its category of models \cite{jiang2023mistral}. These type of LLMs have had success in a large number of NLP tasks, but seem to still lag Masked Language Models in sentence classification. 

\subsubsection{Prompting}

In Causal Modelling, how you prompt the model significantly influences its performance. We tested different prompting strategies in both zero-shot and few-shot settings:

\begin{itemize}
    \item \textbf{Zero-Shot Learning:} Here, we provide the model with a single example and ask it to predict the emotion without any additional references. For zero-shot learning the best prompting technique was: "[INS] Given the following sentence: \verb|{sentence}. ###| Predict which emotion is expressed. Chose one of the following options: neutral, anger, surprise, fear, joy, sadness, disgust, and contempt. Answer in one word only. \verb|###| Answer: [\verb|\INS|]"

    \item \textbf{Few-Shot Learning:} In this setting, we give the model several examples – one for each emotion – along with their corresponding labels. This leverages the model's in-context learning ability, potentially boosting its performance for unseen samples. For few-shot learning the best prompting technique was: "[INS] This is an example of a sad sentence: \verb|{sentence} {repeat for every emotion}|.  \verb|###| Predict the emotion of the following sentence: {sentence}. Chose one of the following options: neutral, anger, surprise, fear, joy, sadness, disgust, and contempt. Answer in one word only. \verb|###| Answer: [\verb|\INS|]"
\end{itemize}

\section{Experimental setup}

\subsection{Data Split Strategy}

We employed a classic data split approach:

\begin{itemize}
\item \label{itm:initial} \textbf{Initial Development}: We combined the training and development sets and shuffled the data. Subsequently, we used 70\% for training, 10\% for validation, and the remaining 20\% as a held-out test set.
\item \label{itm:release} \textbf{Competition Test Set Release}: Upon the competition's test set release, we directly evaluated our models using the platform. To maximize training data, we trained on the combined training set with a 20\% validation split.
\item \label{itm:final} \textbf{Final Model}: Once we selected our best model, we re-trained it on the entire dataset without validation. This re-training didn't yield significant improvements
\end{itemize}

\subsection{Subtask 1}

We'll focus on Subtask 1, where we achieved strong results. The key hyperparameters used:

\begin{itemize}
\item \label{itm:initial} \textbf{Batch Size}: A batch size of 64 provided the best balance. Smaller sizes hurt performance, while larger sizes exceeded our memory constraints.
\item \label{itm:release} \textbf{Fine-Tuning}: We trained for 4 epochs with frozen model weights, followed by 3 epochs with only the last layer unfrozen (as detailed in section \ref{finetunning}).
\item \label{itm:final} \textbf{Classifier}: Our classifier used 128 neurons, 0.5 dropout, and a softmax activation.
\item \label{itm:final} \textbf{Optimization}: We used cross-entropy loss, the AdamW optimizer, and experimented with different learning rates (see section \ref{finetunning}).
\item \label{itm:final} \textbf{Evaluation}: We measured performance using the MulticlassF1Score with 8 classes and 'macro' averaging.
\end{itemize}

\section{Results}

Our top-performing model (\autoref{table:results}) was a fine-tuned FacebookAI/xlm-roberta-large \cite{DBLP:journals/corr/abs-1911-02116}. This highlights the superiority of fine-tuned Masked Language Models (MLMs) over Mistral for sentence classification tasks.  The results suggest that smaller Causal models remain less effective than fine-tuned MLMs in this domain. We also see that few-shot Mistral is worse than zero-shot, probably because too much data in the prompt confuses the model.

\begin{table}[htbp]
\centering
\resizebox{\columnwidth}{!}{
\begin{tabular}{@{}lccc@{}}
\toprule
\textbf{Model} & \textbf{Train} & \textbf{Validation} & \textbf{Test} \\
\midrule
xlm-roberta & 0.74 & 0.57 & \textbf{0.43} \\
mdeberta-v3 & 0.74 & 0.56 & 0.42 \\
bert-multi & 0.66 & 0.48 & 0.35 \\
Mistral zero-shot & - & - & 0.32 \\
Mistral few-shot & - & - & 0.31 \\
distilbert-multi & 0.6 & 0.47 & 0.29 \\

\bottomrule
\end{tabular}%
}
\caption{Results for Subtask 1 - Masked Language Models and Causal Models (Mistral).}
\label{table:results}
\end{table}

In terms of number of epochs, our best model was overfitting when finetuned for too many epochs (\autoref{table:epochs}) and we finally trained for 4 + 3 epochs. 

\begin{table}[htbp]
\centering
\resizebox{\columnwidth}{!}{
\begin{tabular}{@{}lcccc@{}}
\toprule
\textbf{Frozen} & \textbf{Fine-tunning} & \textbf{Training} & \textbf{Validation} & \textbf{Test} \\
\midrule
3 & 2 & 0.67 & 0.53 & 0.4 \\
3 & 3 & 0.71 & 0.55 & 0.42 \\
4 & 3 & 0.74 & 0.57 & \textbf{0.43} \\
4 & 4 & 0.78 & 0.60 & 0.42 \\
4 & 5 & 0.85 & 0.45 & 0.38 \\
5 & 3 & 0.71 & 0.56 & 0.42 \\
\bottomrule
\end{tabular}%
}
\caption{Finding the optimal number of epochs to avoid overfitting. First column contains epochs when training only the classifier. Second columns contains epochs when training the classifier and the last transformer layer.}
\label{table:epochs}
\end{table}

\subsection{Error analysis}

\begin{figure}[htbp]
    \centering
    \includegraphics[width=1\linewidth]{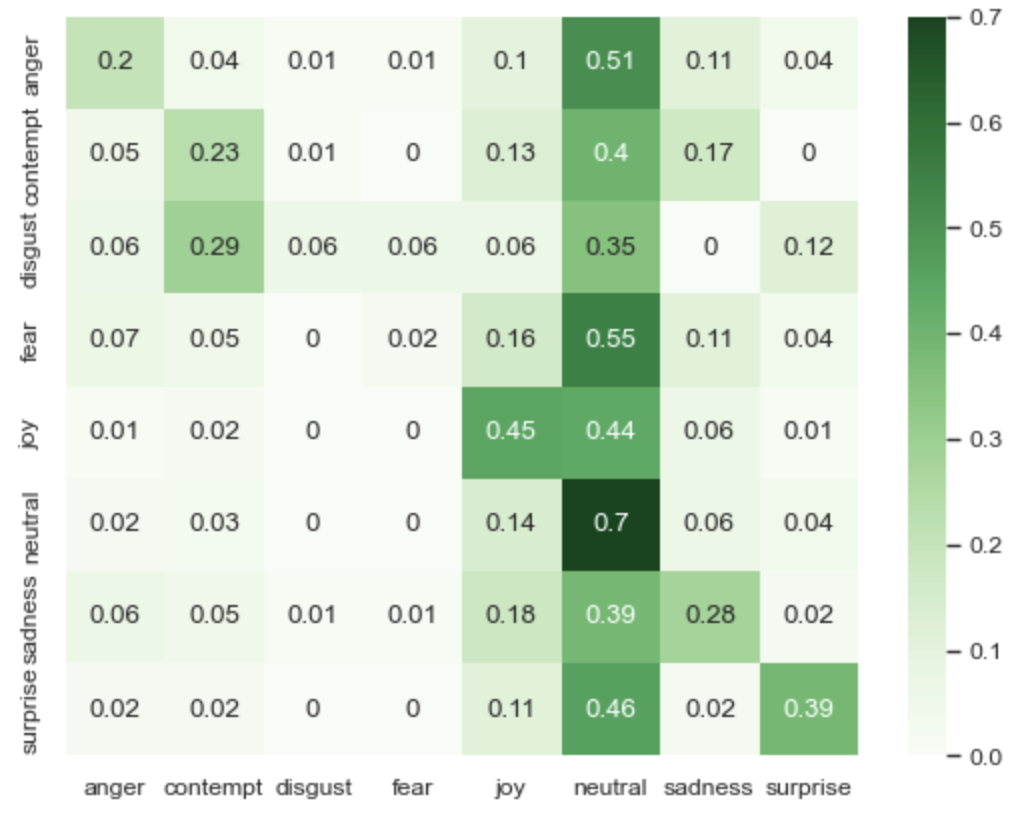}
    \caption{Confusion matrix. On y-axis true labels, on x-axis predicted labels. Values are normalised. }
    \label{fig:confusion_matrix}
\end{figure}

Our confusion matrix (\autoref{fig:confusion_matrix}) reveals that the model overpredicts the 'neutral' emotion, likely due to its prevalence in the training data. This created a bias, leading the model to misclassify instances of other emotions as 'neutral'. While we attempted to mitigate this with class weights in the loss function,  it proves insufficient. In the future, we should explore more robust techniques like oversampling or undersampling to address the class imbalance.

\begin{figure}[htbp]
    \centering
    \includegraphics[width=1\linewidth]{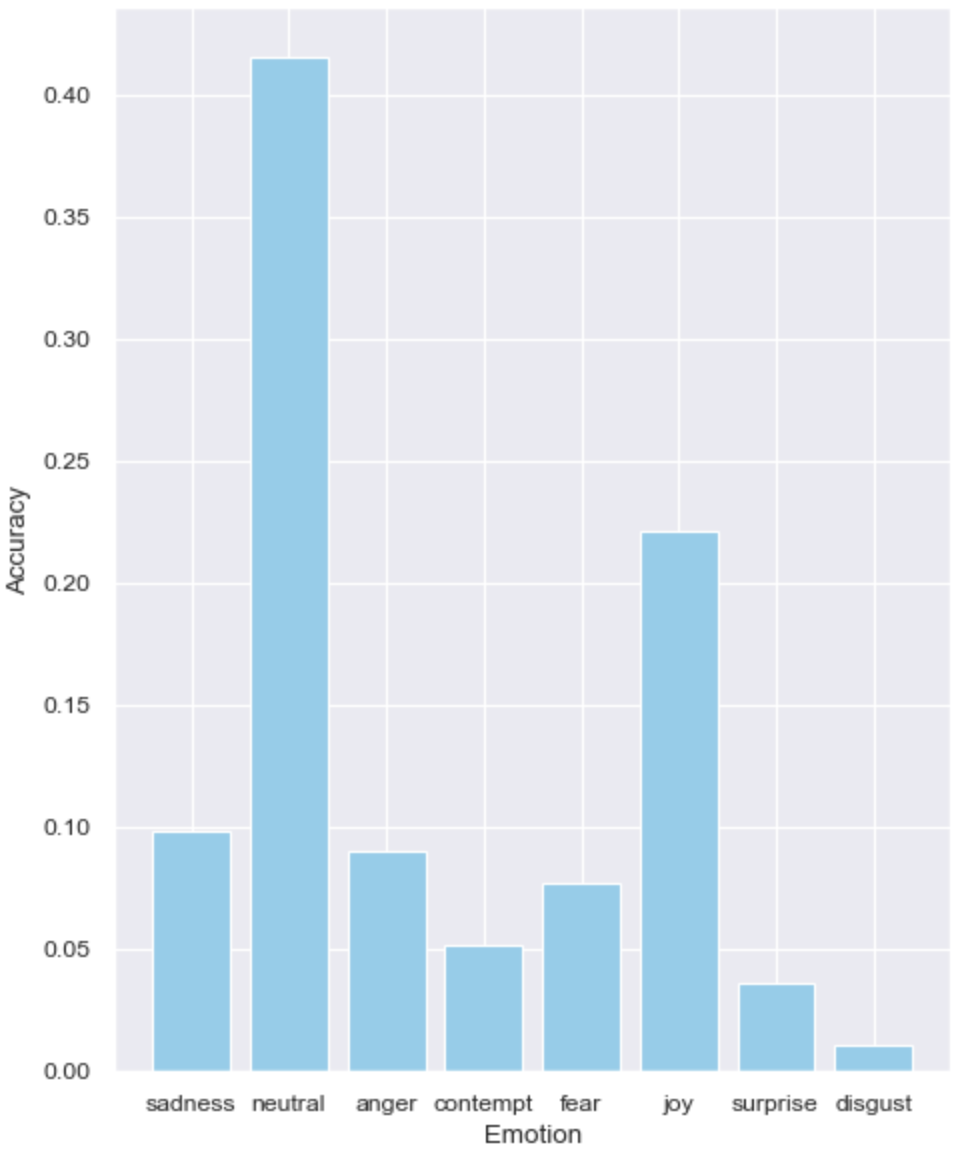}
    \caption{Accuracy by emotion. Accuracy directly correlates with the frequency of each emotion in the training set. }
    \label{fig:accuracy}
\end{figure}

As seen in the emotion accuracy chart (\autoref{fig:accuracy}), the model performs best on the dominant 'neutral' class, along with well-represented emotions like 'joy' and 'sadness'.  Conversely, the model struggles to predict the 'disgust' emotion, which aligns with its under-representation in the training data. This suggests a direct correlation between dataset frequency and model proficiency for each emotion.

\section{Conclusion}

Overall, our system achieved encouraging results in Subtask 1, despite exhibiting some overfitting for dominant labels. While performance on the emotion-flip detection tasks (Subtasks 2 and 3) highlights areas for improvement, we still placed in the first half of the leaderboard. Looking ahead, we plan to investigate hybrid transformer-LSTM architectures for a more nuanced understanding of emotion-flip triggers. Additionally, enriching the data by incorporating a broader conversational context through multi-turn analysis could enhance our model's capabilities. Not least, even though we tried Mistral, there are newer causal models like Mixtral \cite{mixtral2024} and Solar \cite{kim2023solar} which could perform better at this type of task. 

\section*{Acknowledgements}
This work was partially supported by a grant on Machine Reading Comprehension from Accenture Labs and by the POCIDIF project in Action 1.2. ``Romanian Hub for Artificial Intelligence''.

\end{document}